\documentclass{article} 
\usepackage[final]{iclr2023_conference}
\usepackage{times}

\usepackage{amsmath,amsfonts,bm}

\def\eqref#1{equation~\ref{#1}}

\def\1{\bm{1}}

\DeclareMathOperator*{\argmax}{arg\,max}

\DeclareMathAlphabet{\mathsfit}{\encodingdefault}{\sfdefault}{m}{sl}
\SetMathAlphabet{\mathsfit}{bold}{\encodingdefault}{\sfdefault}{bx}{n}

\usepackage[utf8]{inputenc} 
\usepackage[T1]{fontenc}    
\usepackage{hyperref}       
\usepackage{url}            
\usepackage{booktabs}       
\usepackage{amsfonts}       
\usepackage{nicefrac}       
\usepackage{microtype}      
\usepackage{xcolor}         

\usepackage{mathtools} 
\usepackage{booktabs} 
\usepackage{tikz} 

\usepackage{hyperref}
\usepackage{url}

\hypersetup{
    colorlinks=true,
    }
    
\usepackage{subfigure}      
\usepackage{wrapfig}

\usepackage{amsmath}        
\usepackage{amssymb}   
\usepackage{bbm}            
\usepackage{graphicx}       
\usepackage{xcolor}         
\usepackage{subfigure}      
\usepackage{overpic}

\usepackage{multirow}
\usepackage{array}

\usepackage{pgfplots}
\usepackage{mathtools,amssymb}
\usepackage{amsthm}

\newcommand{\norm}[1]{\left\lVert#1\right\rVert}
\usepackage{thmtools, thm-restate}
\usepackage{vcell}
\usepackage{tabularx}
\usepackage[ruled]{algorithm2e}
\makeatletter
\newcommand{\removelatexerror}{\let\@latex@error\@gobble}
\makeatother

\SetCommentSty{mycommfont}
\SetAlFnt{\footnotesize}
\SetAlCapFnt{\footnotesize}
\SetAlCapNameFnt{\footnotesize}

\title{Build generally reusable agent-environment interaction models}

\author{%
  Jun Jin$^{1,}$\thanks{Correspondence to jun.jin1@huawei.com}
  \hspace{1mm}
   Hongming Zhang$^{1,2}$
  \hspace{1mm}
  Jun Luo$^{1}$
\\
\hspace{-5mm}
$^{1}${Huawei Noah's Ark Lab}
  \hspace{1mm}
$^{2}$University of Alberta
  \hspace{1mm}
}

\begin{document}

\maketitle

\begin{abstract}
This paper tackles the problem of how to pre-train a model and make it generally reusable backbones for downstream task learning. In pre-training, we propose a method that builds an agent-environment interaction model by learning domain invariant successor features from the agent's vast experiences covering various tasks, then discretize them into behavior prototypes which result in an embodied set structure. To make the model generally reusable for downstream task learning, we propose (1) embodied feature projection that retains previous knowledge by projecting the new task's observation-action pair to the embodied set structure and (2) projected Bellman updates which add learning plasticity for the new task setting. We provide preliminary results that show downstream task learning based on a pre-trained embodied set structure can handle unseen changes in task objectives, environmental dynamics and sensor modalities.
\end{abstract}

\section{Introduction}
\begin{wrapfigure}{r}{0.35\textwidth}
    \centering
    \includegraphics[width=0.35\textwidth]{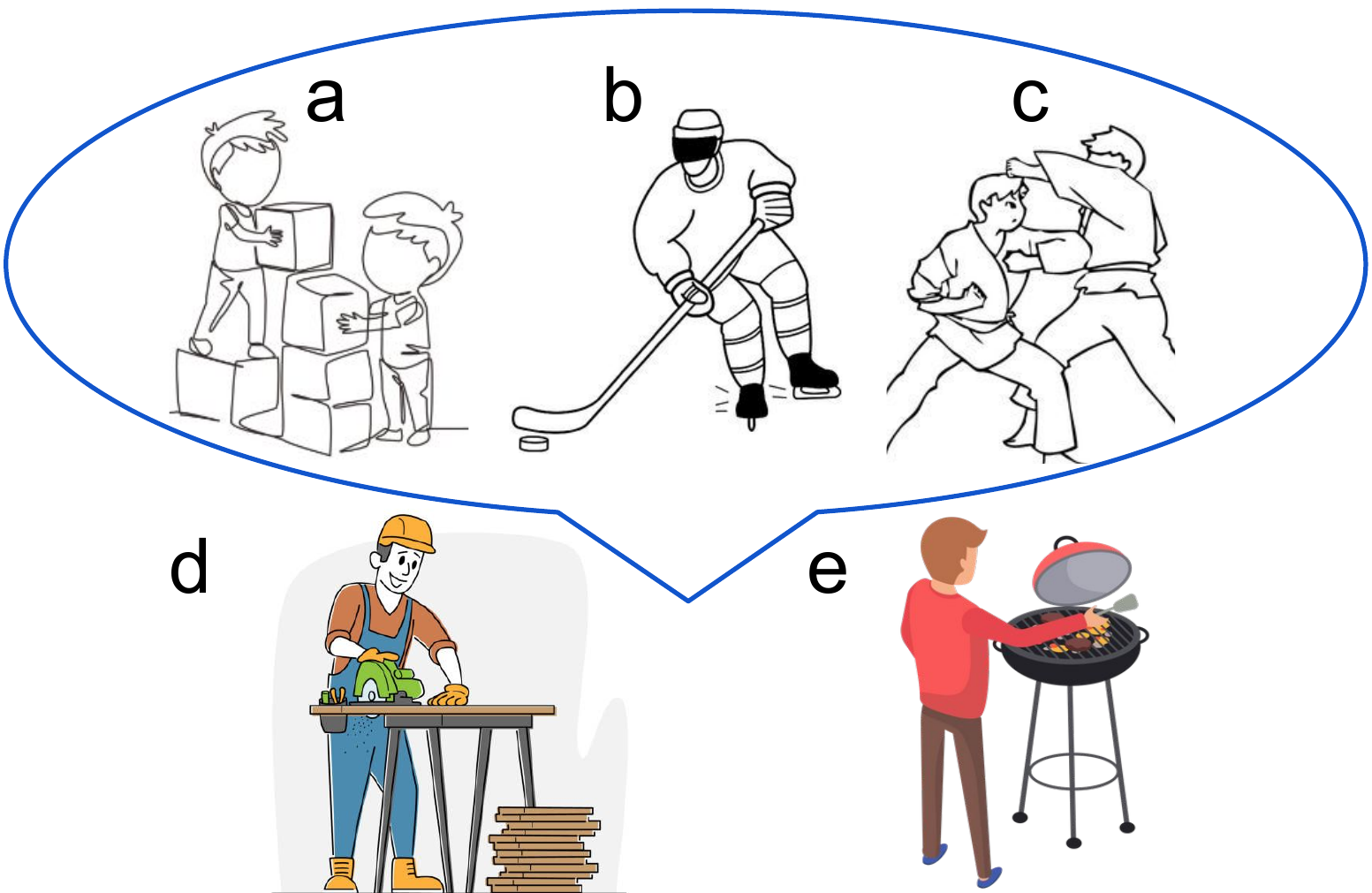}
    \caption{\footnotesize What models are generally shared between heterogeneous tasks that can be reused to accelerate new task learning in unseen settings? \normalsize}
    \vspace{-3mm}
\end{wrapfigure}

Can we pre-train a generally reusable backbone to accelerate downstream task reinforcement learning (RL)? While such backbones are popular in computer vision and natural language processing (e.g., pre-trained ResNet \citep{he2016deep} and pre-trained GPT3 \citep{brown2020language}, pre-trained decision-making backbones that support general downstream task learning are a few. The challenges largely attribute to the vast complexity and variety of decision-making tasks, that the pre-trained model will suffer from the out-of-distribution nightmare when reused to learn a new task. For example, a downstream task's environment dynamics or task objectives may be unseen during pre-training. Or we may need to add an extra sensor modality (e.g., language) to the agent to best fit the task requirement (e.g., understanding human commands). How can we pre-train a backbone that handles the above-unseen changes in downstream task learning, without making any assumptions on the task similarity \citep{duan2016rl,gupta2018meta} of environment similarity \citep{packer2018assessing,rajeswaran2016epopt}? 

Empirically, humans are capable of reusing experience from one task to another, and the tasks we can handle are surprisingly heterogeneous (Fig. 1), from playing blocks, hockey and karate, to learning a carpenter or cooking job. It's natural to ask, (1) what model is generally shared between heterogeneous tasks? And (2) how can it fit in any downstream task learning with unseen settings? 


Systematically, thinking on the above questions results in two perspectives of viewing what component is shared across tasks---the world and the agent \citep{agent}. The first one seeks to build a \textit{Bayesian world model} \citep{ha2018world}, with which, the agent stores prior knowledge about the physical world. For example, ``balls will roll'' and ``walking on ice is slippery''. Then the Bayesian agent selects a proper world configuration for downstream task learning. However, since downstream tasks can be complex and diverse, then how large and how accurate the world model should be is not clear. The second perspective seeks to build an \textit{embedded agency} \citep{orseau2012space}, that the agent gradually constructs knowledge learned from its past experience, respecting the fact that the world is far more complex than the agent. The constructed knowledge can be representations or models and will accelerate downstream task learning. The \textit{embedded agency} perspective can also gradually build a world model for the agent, but unlike the \textit{Bayesian world model}, the agent's past experience shapes what the world model is, how large it is, and how accurate it is. This paper relates to the second viewpoint. 

Though the \textit{embedded agency} view is conceptually appealing, 
technically how to build it and the promising benefits w.r.t. sample efficiency or continual learning capability are not clear. Without being too ambitious to build a complete embedded agency, in this paper, we focus on a more practical starter: \textit{how the agent constructs a model from its embodied experience, and how it accelerates new task learning when environments, task objectives, and even sensor modalities are unseen}. 

Specifically, we format the problem of building a generally reusable model as an embodied set construction approach that constructs a domain invariant set structure using a large dataset covering the agent's vast task experiences. We call it an agent-environment interaction model since the set elements are discretized successor features \citep{barreto2017successor} which represent how a decision-making agent interacts with the world. We call our approach embodied set construction since the set structure is constructed so that it is a shared component across tasks and environments, which can be viewed as an embodiment of the agent. We call it generally reusable since learning by projecting onto the embodied set structure enables learning stability-plasticity. Our contributions are as follows:
\begin{itemize}
    \item We propose an \textit{embodied set construction} approach to pre-train a reusable agent-environment interaction model by firstly learning domain invariant successor features from the agent's past experience, then discretizing them to behaviour prototypes which result in an embodied set structure. Note that the constructed embodied set structure, though optimized to be domain-invariant, still suffers the out-of-distribution (O.O.D.) problem when reused in downstream tasks under unseen settings.
    \item To make the pre-trained model generally reusable, we propose two projection-based techniques that enable learning stability-plasticity: (1) \textit{embodied feature projection} which retains previous knowledge by projecting the new task's observation-action pairs onto the embodied set structure, and (2) \textit{projected Bellman update} which adds plasticity for the new task learning.
    \item Combined together, we propose a complete solution of pre-training a generally reusable model and plugging it as a backbone to accelerate downstream task learning.
\end{itemize}
\begin{figure}[h]
 	\setlength{\belowcaptionskip}{-10pt}
	\centering
	\includegraphics[width=0.8\textwidth]{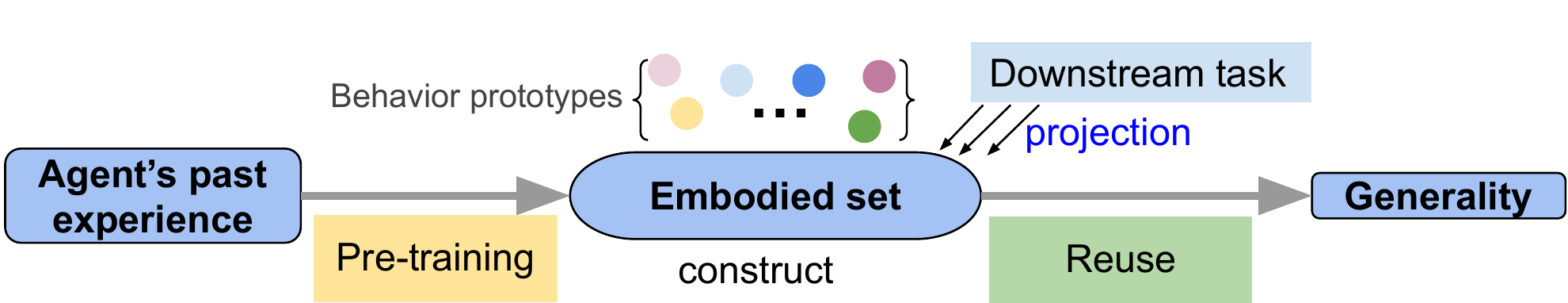}
	\caption{\footnotesize Diagram of our proposed method. \normalsize}
	\label{fig:intro}
\end{figure} 

Our experiments show that downstream task learning based on a pre-trained embodied set structure can handle unseen changes of task objectives, environmental dynamics and sensor modalities, without assuming any environmental or task distribution priors.

\section{Related work}
\label{sec:survey}
Pre-training a backbone to accelerate reinforcement learning is not new. We classify existing works into four categories depending on what is reused in downstream task learning.

(1) \textit{Reusable representations}: Most existing works fall into this category that aims to learn a representation either to improve sample efficiency or task generalization. Methods include task-agnostic ones, for example, using VAE~\citep{kingma2013auto} and a pre-trained ResNet \citep{shah2021rrl} for state representations), self-supervised representation learning methods (RAD \citep{laskin2020reinforcement}, CURL \citep{srinivas2020curl} , CPC ~\citep{oord2018representation}, and ATC \citep{stooke2021decoupling}) for image states, and task-specific ones\citep{Jin2022OfflineLO}, for example, using a behaviour similarity measure (Bi-simulation metrics~\citep{zhang2020learning}) to constrain the representation which results in better generalization performance.

(2) \textit{Reusable policy/value functions}:
This approach aims to pre-train a policy or value function that can be further fine-tuned in downstream task learning. Methods include assuming task distribution priors (meta-RL \citep{duan2016rl}), assuming environmental distribution priors (generalizable RL \citep{packer2018assessing} , robust RL \citep{rajeswaran2016epopt}), and contextual MDP based ones \citep{huang2020one}. Recently, representing a policy using generative sequential models has gained increasing popularity (Gato~\citep{reed2022generalist}, Decision Transformers~\citep{chen2021decision}). However, their capability of downstream task learning is commonly restricted to the training setup, and handling large changes when reused remains challenging~\citep{reed2022generalist}. 

(3) \textit{Reusable skills and options}:
This approach assumes tasks can be decomposed into shareable skills \citep{sutton2011horde} that can be learned via time abstraction---learning options as skills. The difficulty lies in how to discover reusable skills, and it relates to recently proposed works in the \textit{unsupervised RL} literature \citep{laskin2022cic}. Current methods for skill discovery include curiosity and diversity-based ones (intrinsic controls~\citep{gregor2016variational}, DIYAN~\citep{eysenbach2018diversity}, behaviour basis-based ones (Proto-value functions~\citep{mahadevan2005proto}, and variational intrinsic successor features~\citep{hansen2019fast}.









\section{Methodology}
In the below discussions, we use the Partial Markov Decision Process (POMDP) formation  since this setting fits in most real-world reinforcement learning tasks where the full state is not obtainable and the agent is only given observations from its onboard sensors \citep{Igl2018DeepVR}. A POMDP is defined as: $\mathcal{M}=(\mathcal{X}, \mathcal{A}, \mathcal{T}, r, \gamma)$, where  $\mathcal{X}$ denotes the observation space, $\mathcal{A}$ is the action space, $\mathcal{T}$ is the environment transition dynamics, $r$ is the reward function and $\gamma$ denotes the discount factor.

\subsection{Reusable agent-environment interaction models}

\paragraph{(1) Embedded agency: small agent, big world}

Unlike the \textit{Bayesian world model} approach \citep{ha2018world} that aims to construct a big model for the agent as a base knowledge about the world, the \textit{Embedded agency}\citep{orseau2012space} approach takes the perspective on how the small agent perceives, controls and gradually grows based on its own experience. Unlike a \textit{Bayesian world model} where states and models are objective, an \textit{embedded agency} approach is subjective\citep{agent} that agent's past experience will shape its behaviors in new tasks. This idea of intelligence closely relates to Descarte's theory of representationalism and mind-body dualism, and to the modern theory of embodied embedded cognition\citep{haugeland} in the philosophy of mind research. The \textit{embedded agency} approach is appealing in that it promises a scalable learning architecture to build a world model based on the agent's experience in a self-improvement manner, however, it is also ambitious. 
This paper investigates the core idea of \textit{embedded agency} about how an agent's past experience can shape its new task learning in a pre-training and reusing paradigm.  

\paragraph{(2) On agent-environment boundary}
Conceptually, from the \textit{embedded agency} perspective, a generally reusable model in heterogeneous environments and tasks should be the agent itself, since the agent is the only shared component across domains. Practically, using the agent's past experience to extract the agent model is difficult since the agent-environment boundary is not the same as we think in a physical system, like a robot. \citep{Jiang2019OnVF} provide a detailed analysis of this problem. \citep{sutton2018reinforcement} in chapter 3.1 explains that the agent-environment boundary is often task-specific in different abstraction levels and the boundary could change across tasks and environments, which makes exacting a general agent model from its various task data difficult. 

Instead of extracting an agent model, we propose to extract a general agent-environment interaction model that is commonly shared across tasks and environments, and thus can be viewed as an embodiment of the agent. This proposal is explained below.

\paragraph{(3) Generally reusable agent-environment interaction model}
In pre-training, suppose the agent's past experiences $\mathcal{D}$ are generated from M different environments and collected using unlimited behaviour policies induced by unlimited reward functions. This will compose a set of different POMDPs: $\{\mathcal{M}_{i}\}_{i=1}^{M}$, where $\mathcal{M}_{i}=(\mathcal{X}_{i}, \mathcal{A}, \mathcal{T}_{i}, \gamma)$ is a reward-free POMDP that drops the task reward for simplicity. $\mathcal{A}$ without a subscript $i$ means the agent's action space is unchanged across environments since the embedded agency considers the same agent across environments and tasks. Let's assign a one-hot vector encoding $Y_{i}$ as the label for each environment domain $\mathcal{M}_{i}$, then we have a set of domain labels $\mathcal{Y}=\{Y_{i}\}_{i=1}^{M}$, the collected dataset for pre-training can be denoted as $\mathcal{D}=\{D_{Y_{i}}\}_{i=1}^{M}$, where $D_{Y_{i}}$ is the dataset collection in environment domain $Y_{i}$.

We use successor features (SF)\citep{barreto2017successor} to capture the agent-environment interaction model since SFs summarize the dynamics induced by the environment when following a behaviour policy. The first step of building a generally reusable agent-environment interaction model is to learn cross-domain transferable successor features in order to extract a general agent-environment interaction model. For simplicity, we consider a random uniform behaviour policy $\pi_{0}$ for all environment domains. Recall that, SF is defined as the expected cumulative features of $\phi$ by following $\pi_{0}$ starting at a specific state. The SF of $(x^{i}, a)$ in environment domain $Y_{i}$ is defined as:
\begin{equation}
\begin{split}
    \psi^{\pi_{0},i}(x^{{i}}, a)  ={ \mathbb{E}}^{\pi_{0}}_{(x, a, x') \sim \mathcal{T}_{i}} [\sum_{t'=t}^{\infty} \gamma^{t'-t} \phi_{x_{t+1}} |X_{t}=x^{{i}}, A_{t}=a].
\end{split}
\end{equation}
, which satisfies the Bellman Equation as below:
\begin{equation}
\begin{split}
     \psi^{\pi_{0},i}(x^{{i}}, a)  & = \phi_{x^{i}} + \gamma {\mathbb{E}}^{\pi_{0}}_{(x, a, x') \sim \mathcal{T}_{i}} [\psi^{\pi_{0},i}(x^{{i}}_{t+1}, a_{t+1})|X_{t}=x^{{i}}, A_{t}=a]
\end{split}
\label{eq:SF_domain}
\end{equation}
, and it can be learned by minimizing the temporal difference (TD) error:
\begin{equation}
\begin{split}
     \delta_{sf, Y_{i}}^{2} = \norm {\phi_{x^{i}_{t}} + \gamma \psi^{\pi_{0},i}(x^{{i}}_{t+1}, a_{t+1}) - \psi^{\pi_{0},i}(x^{{i}}_{t}, a_{t})}
\end{split}
\label{eq:sf_loss}
\end{equation}
Next, let's consider a set of domains $\mathcal{Y}=\{Y_{i}\}_{i=0}^{M}$. The goal is to learn a successor feature approximation function $f(.;\theta_{sf}): \mathcal{X}_{i} \times \mathcal{A} \rightarrow 
 \boldsymbol {\psi}$ with parameters $\theta_{sf}$ that is transferable across all the domains $\mathcal{Y}$. Inspired by~\citep{feng2019self}, we add two constraints to Eq. \ref{eq:sf_loss} in order to make the learned SF cross-domain transferable, using the mutual information definition $I(.)$:
 \begin{equation}
\begin{split}
   {min} &  \hspace{0.3cm} \mathcal{L}_{sf}  =\frac{1}{M} \sum_{i=1}^{M}\mathbb{E}_{(x, a, x') \sim \mathcal{D}_{Y_{i}}}[\delta_{sf, Y_{i}}^{2}]\\
   s.t. & \hspace{0.3cm}  I(\mathbf{\psi}^{\pi_{0}}, Y) < \epsilon_{u}; \hspace{0.3cm} I(\mathbf{\psi}^{\pi_{0},i}, {x^{i}}) > \epsilon_{l}, \hspace{0.3cm} \forall i \in \{1, ..., M\}
 \end{split}
 \label{eq:total_loss}
\end{equation}
In the first constraint, $\psi^{\pi_{0}}$ is the SF for an arbitrary environment domain, and Y is the corresponding domain label. It limits the mutual information between an SF and its domain label to a threshold $\epsilon_{u}$ in order to make the learned successor feature domain-invariant. The domain index $Y_{i}$ is dropped since it generally applies to all domains. The second constraint term maintains the mutual information between an SF and its input domain-specific observations\footnote{We use the domain index $Y_{i}$ since it requires to estimating a specific domain's marginal distribution $p^{i}(x)$ that $x^{i} \sim p^{i}(x)$ during training, as detailed in Appendix. \ref{sec:app_cross_domain}.} above a threshold $\epsilon_{l}$, in order to prevent the optimization from collapsing to a trivial solution, for example, a random feature space is also domain-invariant but does not include any useful information for task learning. Technical details about optimizing Eq. \ref{eq:total_loss} are shown in Sec. 3.2.

Note that, learning a cross-domain transferable successor feature representation \textit{will not solve} the out-of-distribution (O.O.D.) problem when reusing it in downstream tasks with unseen changes, so that directly plugin the pre-trained successor features, as a common approach in (\citep{barreto2017successor,barreto2020fast}), \textbf{will not make our pre-trained model generally reusable}. We propose two techniques to tackle this problem, \textit{(1) embodied set construction} that discretizes the successor features into prototype sets (Sec. 3.2);  \textit{(2) feature projection} and \textit{projected Bellman updates} to enable learning stability-plasticity (Sec. 3.3). Combined together, we make the pre-trained agent-environment interaction model generally reusable.


\subsection{Pre-training: embodied set construction}
The pre-training process is proposed as an embodied set construction method that has two steps: 
\begin{wrapfigure}[16]{R}{0.43\textwidth} 
   \begingroup
\removelatexerror
\begin{algorithm}[H]
	\SetAlgoLined
	\small
	\KwIn{Offline dataset $\mathcal{D}=\{D_{Y_{i}}\}_{i=1}^{M}$, $\psi^{\pi_{0}}(x, a;\theta_{sf})$, embodied set size $N$}
	\KwResult{Embodied set structure $\Omega^{e}$}
	Initialize an empty embodied set $\Omega^{e}=\{\}$\\
	Initialize an empty successor feature vector list $\mathbf{L}_{sf}=\{\}$\\
	$\mathcal{D} \leftarrow $ Shuffle $(\mathcal{D})$\\
	\For{each $(x, a, x') \in \mathcal{D}$}{
	    \tcp{Compute cross-domain transferable successor features}
	    $\mathbf{L}_{sf} \leftarrow $ Append $\psi^{\pi_{0}}(x, a;\theta_{sf})$  \\
	}
	\tcp{Constructing embodied agent state set}
	K-means clustering ($\mathbf{L}_{sf}, N$)\\
	$\Omega^{e}=\{\mathbf{e}_{i}\}_{i=1}^{N}$ =  cluster-centers as behavior prototypes  \\
	\caption{Embodied Set Construction}
\end{algorithm}
\endgroup
  \end{wrapfigure}

\paragraph{(1) Learn cross-domain transferable successor features}
Let a neural network parameterized with $\theta_{sf}$ to approximate the SF representation: $\psi^{\pi_{0}}(x, a;\theta_{sf})$, where $(x, a)$ comes from an arbitrary environment domain. Our aim is to use loss Eq. \ref{eq:total_loss} to find the optimal $\theta_{sf}$. By adding the Lagrangian multipliers $\lambda_{u}$ and $\lambda_{l}$, the Lagrangian dual of Eq. \ref{eq:total_loss} is:
\begin{equation}
\begin{split}
  \underset{\theta_{sf}}{min}  &  \hspace{0.3cm} \mathcal{L}_{sf} + \lambda_{u} I(\mathbf{\psi}^{\pi_{0}}, {Y}) - \lambda_{l} \sum_{i=1}^{M} I(\mathbf{\psi}^{\pi_{0},i}, {x^{i}}) \\
 \end{split}
 \label{eq:total_loss2}
\end{equation}
Directly optimizing the mutual information (MI) terms of Eq. \ref{eq:total_loss2} in high-dimensional space is challenging, we provide tractable solutions by approximating the upper bound and lower bound, as detailed in Appendix. \ref{sec:app_cross_domain}. Optimizing Eq. \ref{eq:total_loss2} will result in  transferable $\psi^{\pi_{0}}(x, a;\theta_{sf})$.

\paragraph{(2) Discretize to behavior prototypes: construct an embodied set structure}
To facilitate downstream task learning with our proposed \textit{embodied feature projection} and \textit{projected Bellman updates} (Sec. 3.3), we discretize the learned cross-domain successor features in three steps: (i) Compute all the successor features using the learned $\psi^{\pi_{0}}(x, a;\theta_{sf})$ from $\mathcal{D}$ containing all the environment domain samples. (ii) Cluster all the successor features using mini-batch K means. (iii) Collect the center of each cluster to construct a set structure $\Omega^{e}$. This is summarized in Algorithm 1. 

We call $\Omega^{e}$ as \textit{embodied set} since it is the component shared across environments and tasks that can be viewed as the embodiment of the agent. And each $\mathbf{e}_{i} \in \Omega^{e}$ is called a prototype since it is the center of a cluster that represents a prototype agent-environment interaction behaviour.

       

\subsection{Re-use: a backbone for general downstream task learning}

Let a downstream task denoted as $\mathcal{M}_{u}=(\mathcal{X}^{u}, \mathcal{A}, \mathcal{T}^{u}, r^{u}, \gamma)$, where the observation space $\mathcal{X}^{u}$, dynamics $\mathcal{T}^{u}$, and task objectives $r^{u}$ can be unseen in pre-training. The action space $\mathcal{A}$ remains the same since we assume the same agent learning a new task based on its past experience. 

Reusing the pre-trained model under the above heterogeneous settings is difficult due to out-of-distribution concerns. Traditional methods will not work. For example, in methods reusing the pre-trained representation \citep{shah2021rrl,kingma2013auto}, changes in the environment will cause non-stationarity in the observation space that directly plugs in the learned representation will make the learning even worse than learning from scratch. For another example, in methods reusing the pre-trained successor features to fast compute the value functions \citep{barreto2020fast}, changes in task objective or the environment will break the linear reward feature assumption.

We propose to avoid directly plugging in the pre-trained model, but to use the pre-trained model---embodied set $\Omega^{e}$ as a base structure for projection-based techniques to accelerate the downstream task learning by tackling learning stability and plasticity explicitly.

\paragraph{(1) Stability: Retain previous knowledge by embodied feature projection} 
We define a feature projection operator $\Pi_{\Omega^{e}}(x, a)$, which takes the input of $(x, a)$, localizes its projection on the constructed embodied set $\Omega^{e}$, and returns the localized feature vector $\mathbf{e}$:
\begin{equation}
    \Pi_{\Omega^{e}}(x, a)=\mathbf{e}, \hspace{0.2cm} \forall \mathbf{e}\in \Omega^{e}, \text{   find the smallest } \xi(x, a, \mathbf{e})
\end{equation}
, where $\xi(x, a, \mathbf{e}) = \norm{\psi^{\pi_{0}}(x,a;\theta_{sf}) - \mathbf{e}}$, or it can be any other distance metric functions. The feature projection operator $\Pi_{\Omega^{e}}(x, a)$ retains previous knowledge by always matching the new task experience with the closest prototype $\mathbf{e}$ in the embodied set.

Note that $\Pi_{\Omega^{e}}(x, a)$ also works for changed sensor modalities. For example, assuming a task requires the agent to understand textual commands, then adding an extra-textual observation $z$ will augment the observation to $[x, z]$. The feature projection operator $\Pi_{\Omega^{e}}(x, a)$ still applies here by using the unchanged sensory modality part $x$. We will show that in Sec. 5.1 (3).

\paragraph{(2) Plasticity: Adapt to changes by projected Bellman Updates}
We use the below projected Bellman updates to accelerate learning the unknown downstream task $\mathcal{M}_{u}=(\mathcal{X}^{u}, \mathcal{A}, \mathcal{T}^{u}, r^{u}, \gamma)$:
\begin{equation}
    Q^{ \pi}(x,a) = \mathbb{E}_{(x, a, x') \sim \mathcal{T}^{u}}[r^{u} + \gamma V^{ \pi}_{proj}(\Pi_{\Omega^{e}}(x',\underset{a'}{\argmax}Q^{ \pi}(x',a'))]
    \label{eq:bellman1}
\end{equation}
\begin{equation}
   V^{ \pi}_{proj}(\Pi_{\Omega^{e}}(x, a))=\mathbb{E}_{(x, a, x') \sim \mathcal{T}^{u}}[r^{u} + \gamma \underset{a'}{max} Q^{ \pi}(x', a')
   \label{eq:bellman2}
\end{equation}
We maintain two value functions---the task $Q^{ \pi}$ and a projected version $V^{ \pi}_{proj}$, to support each other's learning in a bidirectional improvement manner. The motivation is that the projected function $V^{ \pi}_{proj}$ can learn faster than $Q^{ \pi}$  since it is defined on the pre-trained set $\Omega^{e}$ that retains past experience, but is not accurate since $\Omega^{e}$ does not adapt to the new task setting. Meanwhile, the task Q-value function $Q^{ \pi}$ should be more accurate but will take a longer time if learned from scratch. Learning that alternates Bellman updates Eq. \ref{eq:bellman1} and \ref{eq:bellman2} will play a trade-off between retaining the previous knowledge or adapting to the new tasks.

\paragraph{(3) Reuse example: DQN-embodied}
While the above two techniques \textit{embodied feature projection} and \textit{projected Bellman updates} can generally apply to any RL methods in learning a new task, we use DQN as an example to show how to use them to accelerate downstream task learning. Assume we use a neural network parameterized with $\theta_{u}$ to approximate $Q^{ \pi}$, and another neural network parameterized with $\mathbf{w}_{u}$ to approximate $V^{ \pi}_{proj}$. According to Eq. \ref{eq:bellman1} and \ref{eq:bellman2}, we compute the target value for $Q^{ \pi}$ and $V^{ \pi}_{proj}$ at training iteration i as:
\begin{equation}
\begin{split}
   y_{i} = \mathbb{E}_{(x, a, x') \sim \mathcal{T}^{u}}[r^{u}+\gamma V^{ \pi}_{proj}(\Pi_{\Omega^{e}}(x',\underset{a'}{\argmax}Q(x',a';\theta_{u,i-1}));\mathbf{w}_{u, i-1})]
 \end{split}
 \label{eq:dqn1}
\end{equation}
\begin{equation}
     y_{proj, i} = \mathbb{E}_{(x, a, x') \sim \mathcal{T}^{u}}[r^{u} + \gamma \underset{a'}{max} Q^{ \pi}(x', a';\theta_{u, i-1})]
     \label{eq:dqn2}
\end{equation}
Then, the learning objectives formulated as an LMSE loss can be written as:
\begin{equation}
\begin{split}
  \mathcal{L}_{i}(\theta_{u,i}) = \mathbb{E}_{(x, a, x') \sim \mathcal{T}^{u}}[(y_{i} - Q^{\pi}(x, a;\theta_{u, i}))^{2}] 
 \end{split}
 \label{eq:dqn3}
\end{equation}
\begin{equation}
    \mathcal{L}_{i}(\mathbf{w}_{u,i}) = \mathbb{E}_{(x, a, x') \sim \mathcal{T}^{u}}[( y_{proj, i} - V^{ \pi}_{proj}(\Pi_{\Omega^{e}}(x, a));\mathbf{w}_{u,i})^{2}]
    \label{eq:dqn4}
\end{equation}
During training, we alternate learning $Q^{ \pi}$ and $V^{ \pi}_{proj}$. A full description of the proposed DQN-embodied algorithm is summarized in Appendix \ref{sec:dqn_embodied}.

\section{Experimental setup}
We aim to test \textit{the generality} of our method by increasing the difference w.r.t. the environment, the task objective, and the agent's sensory modality, between pre-training and reuse. 

\textbf{(1) Choosing the environment}\hspace{0.5cm}
To serve the above evaluation purpose, we require the evaluation environment should be easy to change the task objective (reward), environmental dynamics, and sensor modalities. We choose MiniGrid~\citep{minigrid} since it is highly re-configurable and easy to reproduce results. We test tasks including ``goal reaching'',  ``picking up the grey box'' and  ``putting the green key near the yellow ball''. 

 We use the default partial observability setting that the agent can only observe nearby $7\times7$ cells. Each cell is described with 3 input values: [OBJECT\_IDX, COLOR\_IDX, STATE], which gives the object id, cell color and door state (open, closed, locked). Therefore, the observation $x$ is a $7\times7\times3$ dimensional vector. In changing the sensor modality test, we manually add a textual observation by using the embedding of a text $z_{t}$ to augment $x_{t}$ with $[x_{t}, z_{t}]$.

The agent's action space remains unchanged during pre-training and downstream task learning, which is a discrete set containing 6 action choices: turn left, turn right, move forward, pick up an object, drop the object being carried, and toggle (open doors, interact with objects).

\textbf{(2) Dataset collection}\hspace{0.5cm}
We consider 15 different room layouts environments during pre-training, specifically, 1 layout using 1 room, 4 layouts using 2 rooms and 10 layouts using 4 rooms. The room layouts are randomly generated. To fully utilize the agent's action capability, especially for ``picking up an object'', ''dropping an object being carried'', and ``toggle'', we randomly place objects (blue keys, blue doors, and different colored balls). For each room layout, we randomly configure 10 different object placement scenarios. Therefore, we have 150 environments to generate the agent's experience for pre-training. We then let the agent explore uniformly to navigate the rooms and interact with the objects without any task goals to collect samples. Therefore, we have 150 task-free POMDPs $\{\mathcal{M}_{i}\}_{i=1}^{150}$, which result in 150 environment domains $\{Y_{i}\}_{i=1}^{150}$, with each $Y_{i}$ is a one-hot encoding vector to denote the domain label. 

In summary, during data collection, we created 150 play zones for the agent to explore freely without defining any task goals. For each domain, we collected $1M$ samples which compose the dataset $\mathcal{D}=\{\mathcal{D}_{i}\}_{i}^{150}$ with $150M$ samples. Each samples is a tuple $(x, a, x')$.


\textbf{(3) Pre-training: embodied set construction}\hspace{0.5cm}
Following~\citep{Hoang2021SuccessorFL}, we pre-train a variational auto-encoder (VAE~\citep{kingma2013auto}) that maps observation $x_{t}$ to a feature vector $\phi_{t}$, which is then used to learn the cross-domain transferable successor features by learning the function $\psi^{\pi_{0}}(x, a;\theta_{sf})$ using methods proposed in Sec. 3.2. During embodied set construction, we set $N=10K$ to discretizes all SFs into 10K clusters and construct the embodied set $\Omega^{e}$ representing 10K prototypes behaviors.

\textbf{(4)Re-use: a backbone for downstream task learning}\hspace{0.5cm}
Given the pre-trained embodied set structure $\Omega^{e}$, we test how it can be reused to accelerate downstream task learning using our proposed DQN-embodied method (see Algorithm 2), as described below.

\section{Preliminary results}

\begin{figure*}[t]
 	\setlength{\belowcaptionskip}{-10pt}
	\centering
	\includegraphics[width=0.8\textwidth]{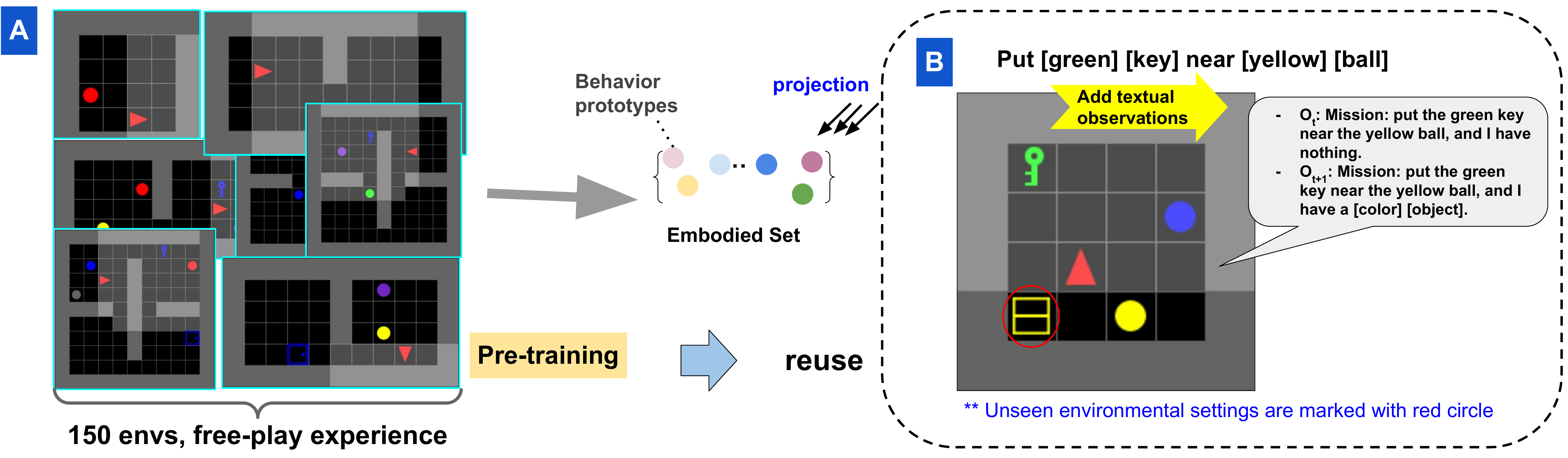}
	\caption{\footnotesize \textbf{A:} Environments for collecting the pre-training dataset with different room layouts and random objects (blue keys, blue doors, coloured balls) for the agent to interact with. \textbf{B:} Downstream task ``putting the green key near the yellow ball'' uses an unseen environment (green key, coloured boxes unseen during pre-training), unseen task objective, and unseen sensor modality by adding textual observations. Note this task requires text prompts otherwise it is difficult to solve by a vanilla RL agent\citep{minigrid} \normalsize}
     \vspace{1.0mm}
	\label{fig:exp_setup}
\end{figure*}

We conduct the \textit{generality test} by adding differences between pre-training and reuse. Fig. \ref{fig:exp_setup} gives an overview of task setup in the below three tests. 




\textbf{Unseen task objective + unseen environment + unseen sensor modalities}\hspace{0.5cm}
We test the generality of our proposed method by adding an unseen task objective, environment settings, and sensor modality in downstream task learning. The task is ``putting the green key near the yellow ball'' wherein the agent needs to understand ``green key'' (unseen in pre-training), ``yellow ball'' and ``near''. The task requires the agent to find the green key, carry it and drop the key in a cell nearby the yellow ball. The reward function is defined as $r(x, a, x')=0.1$ (if pick up the green key), 10.0 (if succeed).

\textbf{Adding textual sensor modality:} Note that this task requires text prompts otherwise is difficult to solve. We manually design three types of textual observations for the agent--- The mission statement ``Put the green key near the yellow ball'' + different textual statuses: \{``and I have nothing'', ``and I have the [color] [object]''\} depends on what the agent is carrying. The textual observation it then converted to an embedding vector $o$ using a pre-trained sentence transformer MiniLM (384 dimensional output, 80 MB model size) \citep{Wang2020MiniLMDS} using a libary provided by \citep{reimers-2019-sentence-bert}. Therefore, the agent's observation at time t is augmented as : $[x_{t}, o_{t}]$, where $x_{t} \in \mathbb{R}^{7\times 7\times 3}$ is the same observation as in pre-training, and $o_{t} \in \mathbb{R}^{384}$ is the extra sensor modality added to the agent to provide textual observations.  

\textbf{Baselines:} We compare our method (\textit{DQN-embodied}) to the following baselines. (1) Without pre-training: vanilla DQN~\citep{Mnih2013PlayingAW}; (2) Reuse the pre-training dataset: experience replay (ER) which reuses the whole dataset $\mathcal{D}$ in the buffer to alleviate the forgetting issue~\citep{Rolnick2019ExperienceRF}; (3) Reuse a pre-trained backbone: DQN-VAE~\citep{kingma2013auto} is used as a backbone for downstream task learning; (4) Without textual observations (DQN-no-text) using a vanilla DQN~\citep{Mnih2013PlayingAW}. Note that baselines DQN-VAE can not directly handle an extra sensor modality. When taking the $o_{t}$ as inputs to the above baselines, we concatenate the VAE encoded observation with $o_{t}$. 
\begin{wrapfigure}{r}{0.5\textwidth}
    \centering
		\subfigure{\includegraphics[width=0.5\textwidth]{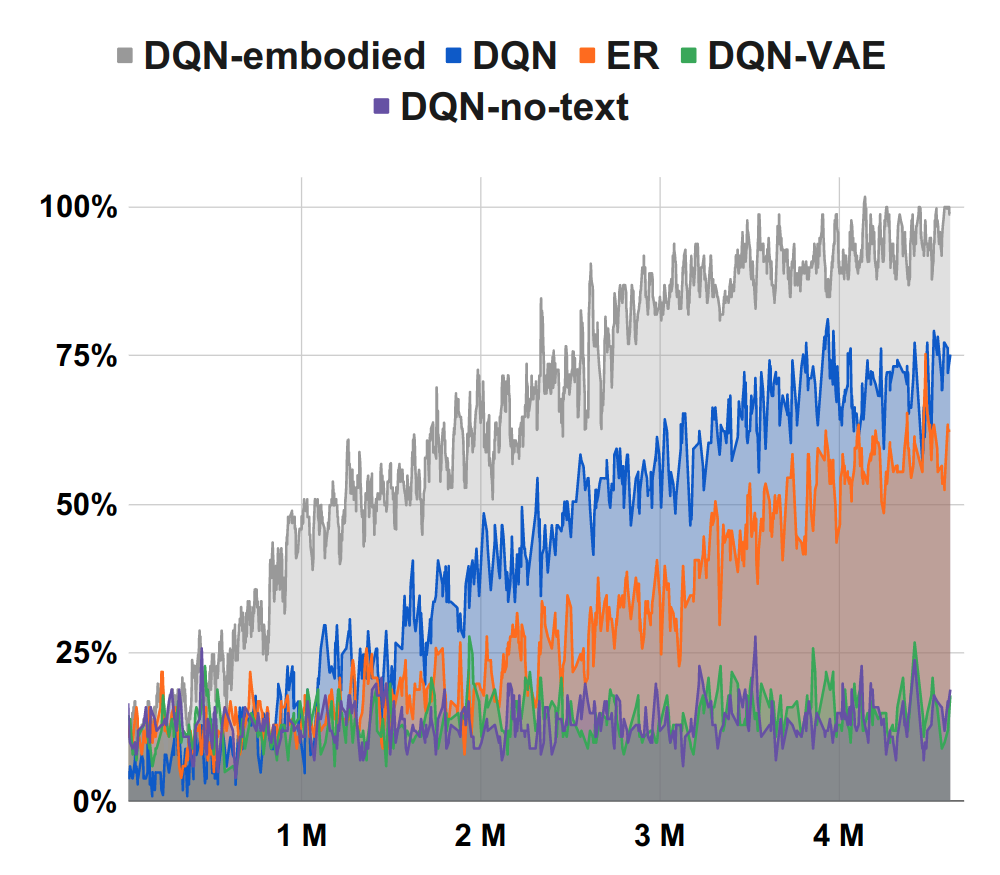}}
    \caption{\footnotesize Preliminary results of reusing generality test. X-axis represents the sample size. Y-axis represents the success rate of 100 independent runs.\normalsize} 
    \label{fig:gen_curve}
   \vspace{-4.5mm}
\end{wrapfigure}

We test the downstream task learning performance of all methods by evaluating the success rate every 10K training steps using 100 independent runs with different random seeds. Results (Fig. \ref{fig:gen_curve}) show that (1) Without textual observations, a vanilla DQN agent can not solve the task; (2) Since the new task setting is heterogeneous compared to tasks in the training pools, directly reusing previous experience (the ER baseline) will deteriorate the downstream task learning performance (vanilla DQN is better than ER); (3) Also, since the new task is never seen during pre-training, DQN based on pre-trained VAE performs the worst. (4) For the test task that has unseen task objectives (put a color object near another color object), unseen environmental settings (box object never seen in pre-training), and unseen sensor modalities (adding textual observations), projecting the downstream task samples on the pre-trained embodied set will result in better sample efficiency than all baselines.



\section{Conclusion}
This paper tackles the problem of pre-training a generally reusable model for downstream tasks with unseen task objectives, environments, and even with unseen sensor modalities. Our approach is inspired by the \textit{embedded agency} perspective that the agent's past experience shapes its future decision-making behaviors. We propose to learn an agent-environment interaction model by extracting cross-domain transferable successor features using the agent's various task experiences. To make the model generally reusable, we propose to discrete it into behavior prototypes, resulting in an embodied set structure that further supports \textit{embodied feature projection} and \textit{projected Bellman updates} to enable downstream task learning with heterogeneous settings. Preliminary results show that downstream task learning by projecting on our proposed embodied set structure can handle unseen changes in task objectives, environmental dynamics and sensor modalities.

For future works, more comprehensive experiments are needed to understand why and how the pre-trained embodied set structure can be generally reused in downstream task learning. It's also worth working on methods that can continually update the constructed \textit{embodied set structure} given that more task experience is gained, thus forming an \textit{embedded agency} version of continual reinforcement learning~\citep{Khetarpal2020TowardsCR} solution.

\bibliography{iclr2023_conference}
\bibliographystyle{iclr2023_conference}

\appendix
\section{Appendix}
\subsection{Learn cross-domain transferable successor features}
\label{sec:app_cross_domain}
Assume a neural network with parameters $\theta_{sf}$ is used to approximate the SF representation denoted as $\psi^{\pi_{0}}(x, a;\theta_{sf})$, where $(x, a)$ comes from an arbitrary environment domain. Our aim is to use loss Eq. \ref{eq:total_loss} to find the optimal $\theta_{sf}$. By adding the Lagrangian multipliers $\lambda_{u}$ and $\lambda_{l}$, the Lagrangian dual of Eq. \ref{eq:total_loss} is:
\begin{equation}
\begin{split}
  \underset{\theta_{sf}}{min}  &  \hspace{0.3cm} \mathcal{L}_{sf} + \lambda_{u} I(\mathbf{\psi}^{\pi_{0}}, {Y}) - \lambda_{l} \sum_{i=1}^{M} I(\mathbf{\psi}^{\pi_{0},i}, {x^{i}}) \\
 \end{split}
 \label{eq:total_loss2}
\end{equation}
Directly optimizing the mutual information (MI) terms of Eq. \ref{eq:total_loss2} in high-dimensional space is challenging, we provide tractable solutions by approximating the upper bound and lower bound.

\paragraph{Upper bound of $I(\mathbf{\psi}^{\pi_{0}}, {Y})$}
Following~\citep{feng2019self}, the $I(\mathbf{\psi}^{\pi_{0}}, {Y})$ is upper bounded by replacing one of the marginal distributions with a variational posterio distribution. For any distribution $q(Y)$, we can have an upper bound of  $I(\mathbf{\psi}^{\pi_{0}}, {Y})$:
\begin{equation}
\begin{split}
  I(\mathbf{\psi}^{\pi_{0}}, {Y}) & = \mathbb{E}_{p_{ }(\mathbf{\psi}^{\pi_{0}}, Y)}[log\hspace{0.1cm}{p_{ }}(Y|\psi^{\pi_{0}}) - log \hspace{0.1cm} p(Y)] \\
  & = \mathbb{E}_{p_{ }(\mathbf{\psi}^{\pi_{0}})}[D_{KL}(p_{ })((Y|\psi^{\pi_{0}})||q(Y)) - D_{KL}(p(Y)||q(Y))] \\
  & \leq \mathbb{E}_{p_{ }(\mathbf{\psi}^{\pi_{0}})}D_{KL}(p_{ }(Y|\psi^{\pi_{0}})||q(Y))\\
 \end{split}
\end{equation}
, where $q(Y)$ can be empirically estimated in a non-parametric way, like using kernel density estimation from the dataset. Therefore, minimization of the KL-divergence term will result in minimizing the mutual information term, which can be formulated as an adversarial training objective, as derived in \citep{feng2019self}:
\begin{equation}
\begin{split}
  \underset{\theta_{sf}}{min} \hspace{0.1cm} \underset{\theta_{u}}{max} \mathcal{L}_{u} = \mathbb{E}_{p_{\theta_{sf}}(\psi^{\pi_{0}}, Y)}[log \hspace{0.1cm} q_{\theta_{u}}(Y|\psi^{\pi_{0}}) - log \hspace{0.1cm} q(Y)]
 \end{split}
\end{equation}
, where $q_{\theta_{u}}(Y|\psi^{\pi_{0}})$ is a classifier that predicts the probability of the sample belonging to label $Y$, and $q(Y)$ is a constant that can be dropped during optimization.

\paragraph{Lower bound of $I(\mathbf{\psi}^{\pi_{0},i}, {x^{i}})$}
There are multiple ways (~\citep{oord2018representation,nowozin2016f}) to approximate a lower bound of the mutual information term. We follow~\citep{feng2019self} that maximizes the MI term by maximizing the Jensen-Shannon divergence form of MI, $\hat{I}^{JSD}(\mathbf{\psi}^{\pi_{0},i}, {x^{i}};\theta_{l})$, where $\theta_{l}$ parameterizes the JSD. Readers can refer to~\citep{nowozin2016f,feng2019self} for more details. 

A complete loss term of Eq. \ref{eq:total_loss2} can be rewritten as a minimax optimization objective:
\begin{equation}
\begin{split}
  \underset{\theta_{sf},\theta_{l}}{min}  \hspace{0.1cm} \underset{\theta_{u}}{max} \mathcal{L}_{sf} + \lambda_{u} \mathcal{L}_{u} - \lambda_{l} \sum_{i=1}^{M} \hat{I}^{JSD}(\mathbf{\psi}^{\pi_{0},i}, {x^{i}};\theta_{l})
 \end{split}
\end{equation}
, which is now tractable.

\subsection{DQN-embodied}
\label{sec:dqn_embodied}
By using the projected Bellman updates (eq. \ref{eq:bellman1} and eq. \ref{eq:bellman2}) proposed here, we compute the target values as described in eq. \ref{eq:dqn1} and eq. \ref{eq:dqn2}. Then, the full DQN-embodied algorithm alternates optimizing $Q^{ \pi}$ and $V^{ \pi}_{proj}$ is as below:

\begingroup
\removelatexerror
\begin{algorithm}[H]
	\SetAlgoLined
	\small
        Given pre-trained embodied set $\Omega^{e}$, feature projection operator $\Pi_{\Omega^{e}}(x,a)$\\
	Initialize $Q^{ \pi}(.;\theta_{u})$, $V^{ \pi}_{proj}(.;\mathbf{w}_{u})$,  and replay buffer $\mathcal{D}$\\
	\For{i=1:N}{
	    \tcp{Replay buffer}
	    \For{t=0:T}{
	        $\epsilon$ greedy select action $a_{t}$ based on ${max}_{a} Q^{ \pi}(x_{t},a_{t};\theta_{u,i})$\\
	        Execute action $a_{t}$ in environment, observe $x_{t+1}, r_{t}$\\
	        Append transition sample $(x_{t}, a_{t}, r_{t}, x_{t+1})$ in $\mathcal{D}$\\
	        Randomly sample batch transitions $\mathcal{B}=\{(x, a, r, x')\}$ from $\mathcal{D}$\\
	        \tcp{Learn $Q^{ \pi}$}
	        Set $y_{i} = \mathbb{E}_{(x, a, x') \sim \mathcal{T}^{u}}[r^{u}+\gamma V^{ \pi}_{proj}(\Pi_{\Omega^{e}}(x',\underset{a'}{\argmax}Q(x',a';\theta_{u,i-1}));\mathbf{w}_{u, i-1})]$\\
	        Perform gradient descent step on $\mathcal{L}_{i}(\theta_{u,i})=(y_{i} - Q^{ \pi}(x, a;\theta_{u, i}))^{2}$\\
	        \tcp{Learn $V^{ \pi}_{proj}$ }
	        Set $ y_{proj, i} = \mathbb{E}_{(x, a, x') \sim \mathcal{T}^{u}}[r^{u} + \gamma \underset{a'}{max} Q^{ \pi}(x', a';\theta_{u, i-1})]$\\
	        Perform gradient descent step on $\mathcal{L}_{i}(\mathbf{w}_{u,i})=( y_{proj, i} - V^{ \pi}_{proj}(\Pi_{\Omega^{e}}(x, a));\mathbf{w}_{u,i})^{2}$
	    }
	}
	\caption{DQN-embodied}
\end{algorithm}
\endgroup

\end{document}